\documentclass[letterpaper, 10 pt, conference]{ieeeconf}
\usepackage{cuted}
\usepackage{booktabs}
\usepackage{multirow}
\usepackage[dvipsnames]{xcolor}

\IEEEoverridecommandlockouts                             
\usepackage{url}
\usepackage{color}
\usepackage{wrapfig}
\usepackage{caption}
\usepackage{multirow}
\usepackage{ragged2e}
\usepackage{subcaption} 
\usepackage{tabularx}
\usepackage{algorithm}
\usepackage{bm} 

\usepackage{amsmath}
\usepackage{float}
\usepackage{setspace}
\usepackage{xcolor,pifont}
\usepackage{graphicx}
\usepackage{mathrsfs}
\usepackage{siunitx}
\usepackage{amssymb}
\usepackage{mwe} 
\usepackage{nicefrac}
\usepackage{algorithm}
\usepackage[noend]{algpseudocode}

\makeatletter
\newcommand\fs@spaceruled{\def\@fs@cfont{\bfseries}\let\@fs@capt\floatc@ruled
  \def\@fs@pre{\vspace{0.4\baselineskip}\hrule height.8pt depth0pt \kern2pt}%
  \def\@fs@post{\vspace{-0.4\baselineskip}\kern2pt\hrule\relax\vspace{-12pt}}%
  \def\@fs@mid{\kern2pt\hrule\kern2pt}%
  \let\@fs@iftopcapt\iftrue}
\let\NAT@parse\undefined
\makeatother

\usepackage{tikz}
\usepackage{pgfplots}
\pgfplotsset{compat=1.18}
\usepgfplotslibrary{colorbrewer}
\pgfplotscreateplotcyclelist{linestyles}{
    {blue, thick},
    {red, dashed},
    {green, dotted}
}
\usepackage{booktabs}
\usepackage{pifont}

\usepackage{listings}
\lstnewenvironment{itemlisting}[1][]
 {%
  \mbox{}
  \vspace*{-\baselineskip}
  \lstset{
    xleftmargin=\leftmargin,
    linewidth=\linewidth,
    #1
  }%
 }
 {}
\usepackage{flushend}
\usepackage{multicol}

\usepackage{hyperref}
\hypersetup{
     colorlinks   = true,
     linkcolor    = blue,
     citecolor    = red
}

\newcommand{\cmark}{\ding{51}}%
\newcommand{\xmark}{\ding{55}}%
\newcommand{\greencheck}{{\color{green}\cmark}}
\newcommand{\redcross}{{\color{red}\xmark}}

\usepackage{lipsum}

\usepackage{arydshln}
\title{\LARGE \bf TAG-K: Tail-Averaged Greedy Kaczmarz for Computationally Efficient and Performant Online Inertial Parameter Estimation}

\author{Shuo Sha$^{1}$, Anupam Bhakta$^{2*}$, Zhenyuan Jiang$^{1*}$, Kevin Qiu$^{1*}$,\\Ishaan Mahajan$^{1}$, Gabriel Bravo-Palacios$^{3,4}$, Brian Plancher$^{3,4}$%
\thanks{$^{1}$ School of Applied Science, Columbia University.}
\thanks{$^{2}$ Columbia College, Columbia University.}%
\thanks{$^{3,4}$ Barnard College, Columbia University and Dartmouth College.}%
\thanks{$^*$ These authors contributed equally.}
\thanks{Corresponding Author Contact: {\tt\footnotesize plancher@dartmouth.edu}}}

\begin{document}
\maketitle
\thispagestyle{empty}
\pagestyle{empty}


\begin{abstract}
    Accurate online inertial parameter estimation is essential for adaptive robotic control, enabling real-time adjustment to payload changes, environmental interactions, and system wear. Traditional methods often struggle to track abrupt parameter shifts or incur high computational costs, limiting their effectiveness in dynamic environments and for computationally constrained robotic systems. We introduce TAG-K, a lightweight extension of the Kaczmarz method that combines greedy randomized row selection for rapid convergence with tail averaging for robustness under noise and inconsistency. This design enables fast, stable parameter adaptation while retaining the low per-iteration complexity inherent to the Kaczmarz framework.
We evaluate TAG-K in synthetic benchmarks and quadrotor tracking tasks against RLS, KF, and other Kaczmarz variants. TAG-K achieves 1.5×–1.9× faster solve times on laptop-class CPUs and 4.8×–20.7× faster solve times on embedded microcontrollers. More importantly, these speedups are paired with improved robustness to measurement noise and a 25\% reduction in estimation error, leading to nearly 2× better end-to-end tracking performance.
Website, documentation, and code available at: \url{https://a2r-lab.org/TAG-K/}.
\end{abstract}

\section{Introduction} \label{sec:intro}
Model-based control has enabled significant advances across humanoids, quadrupeds, and aerial robots~\cite{Wensing_etal_OBC_24,atlas,taskable_agility,kim2019highly,shin2022design,Li_Wensing_CAFE_2025,krinner2024mpccmodelpredictivecontouring}. A prerequisite for such precise prediction and stable control is accurate \textit{system identification} (SysID), i.e., determining both the structure and parameters of the dynamics model. While the model structure is often known, inertial parameters such as masses, centers of mass, and inertia tensors vary during operation due to payloads, contacts, or reconfiguration.  
\emph{Online parameter estimation} addresses this challenge by updating parameters in real time, allowing robots to adapt their dynamics models to changing conditions, enabling advances in manipulation~\cite{zhang2025provablysafeonlineidentification, baek2024onlinelearningbasedinertialparameter}, adaptive trajectory tracking~\cite{LOPEZSANCHEZ2021243,electronics13020347}, and aerial robotics~\cite{aerial_param_est,LOPEZSANCHEZ2021243}.  

However, traditional online estimators face fundamental limitations. Recursive Least Squares (RLS) efficiently handles measurement noise and can accommodate slowly varying parameters through a forgetting factor~\cite{plackett_rls,sayed2003,JIANG2004403,rls_variable_forgetting}, but adapts poorly to abrupt parameter shifts. The Kalman Filter (KF), as a more general state-space estimator, explicitly models both process and measurement noise but struggles with time-varying process noise (e.g., unmodeled disturbances) and its updates are more computationally expensive~\cite{kalman1960new,kf_textbook,PlancherAppliedApproxMM}. 
Learning-based approaches provide a data-driven alternative by either fitting the dynamics model end-to-end or embedding physics constraints/parameters during training~\cite{PINN_Review, LJUNG20201175, PINNSysiD}. While these methods can support high-order complex nonlinear systems, like deformable or soft-body objects \cite{deformable_learning_dynamics, particle_grid_dynamics}, they may not always offer consistent and reliable physics-based guarantees for systems with well-established models~\cite{lutter2020differentiablephysicsmodelsrealworld,qu2020combiningmodelbasedmodelfreemethods}. Across all of these methods, the computational complexity of these approaches limits their applicability, particularly for embedded, compute-limited systems.

This motivates the development of lightweight, adaptive, and robust alternatives. One promising direction is to enhance classical methods with improved underlying solvers, such as the Kaczmarz family of algorithms~\cite{karczmarz1937angenaherte,rk,Zouzias_2013,chen2018kaczmarz}. These methods solve linear systems by sequentially assimilating relevant measurements through projections onto valid manifolds, enabling computationally efficient solutions without requiring access to the entire data matrix~\cite{bai2018greedy}, and have found recent success in robotics applications~\cite{kaczmarz_robotics}.

As such, we introduce the \emph{Tail-Averaged Greedy Randomized Kaczmarz (TAG-K)} algorithm for online parameter estimation in robotic systems. TAG-K leverages the Kaczmarz family’s efficiency in projecting onto subsets of system manifolds and augments it with greedy row selection for rapid closed-loop adaptation and tail averaging for robustness to noise. This yields a high-performance framework well-suited for online estimation on embedded robotic hardware.

We evaluate TAG-K on challenging quadrotor tracking tasks, where it achieves 1.5$\times$-1.9$\times$ faster solve times on laptop-class CPUs, and more importantly, improved resilience to measurement noise, and a 25\% reduction in estimation error, leading to nearly 2$\times$ better end-to-end tracking performance. 
To further validate our computational efficiency, and demonstrate suitability for embedded deployment, we benchmark TAG-K on synthetic parameter estimation tasks on an embedded microcontroller. TAG-K achieves 4.8$\times$–20.7$\times$ faster solves, highlighting the importance of light-weight online estimation algorithms for systems with limited processing power.
\section{Background and Related Work} \label{sec:related}
\subsection{Online Parameter Estimation}

Consider a robot with $n_d$ degrees of freedom and $n_b$ rigid bodies. Its dynamics follow the Newton–Euler equations: 
\begin{equation} 
\mathbf{M}(\boldsymbol{\theta}, \mathbf{q}) \ddot{\mathbf{q}} + \mathbf{C}(\boldsymbol{\theta}, \mathbf{q}, \dot{\mathbf{q}}) \dot{\mathbf{q}} + \mathbf{g}(\boldsymbol{\theta}, \mathbf{q}) = \mathbf{F}, 
\label{N-E dynamics} 
\end{equation} 
where $\boldsymbol{\theta} \in \mathbb{R}^{10n_b}$ is the inertial parameters for each rigid body (the mass and $3\times3$ inertia-matrix components), $\mathbf{q} \in \mathbb{R}^{n_d}$ is the joint-space configuration, $\mathbf{M}, \mathbf{C} \in \mathbb{R}^{n_d \times n_d}$ are the mass and Coriolis force matrices, and $\mathbf{g}, \mathbf{F} \in \mathbb{R}^{n_d}$ are the gravity and net force vectors.

A key property is that Eq.~\eqref{N-E dynamics} is \emph{linear} in the inertial parameters, giving the regression model: \begin{equation} \mathbf{A}(\mathbf{q}, \dot{\mathbf{q}}, \ddot{\mathbf{q}})\,\boldsymbol{\theta} = \mathbf{F}, \label{measurement model} \end{equation} where $\mathbf{A}$ is the regressor matrix derived from kinematic measurements~\cite{sys_id_1985,measuring_A_matrix}. 

\begin{figure}[!t]
    \centering
    \includegraphics[width=\linewidth]{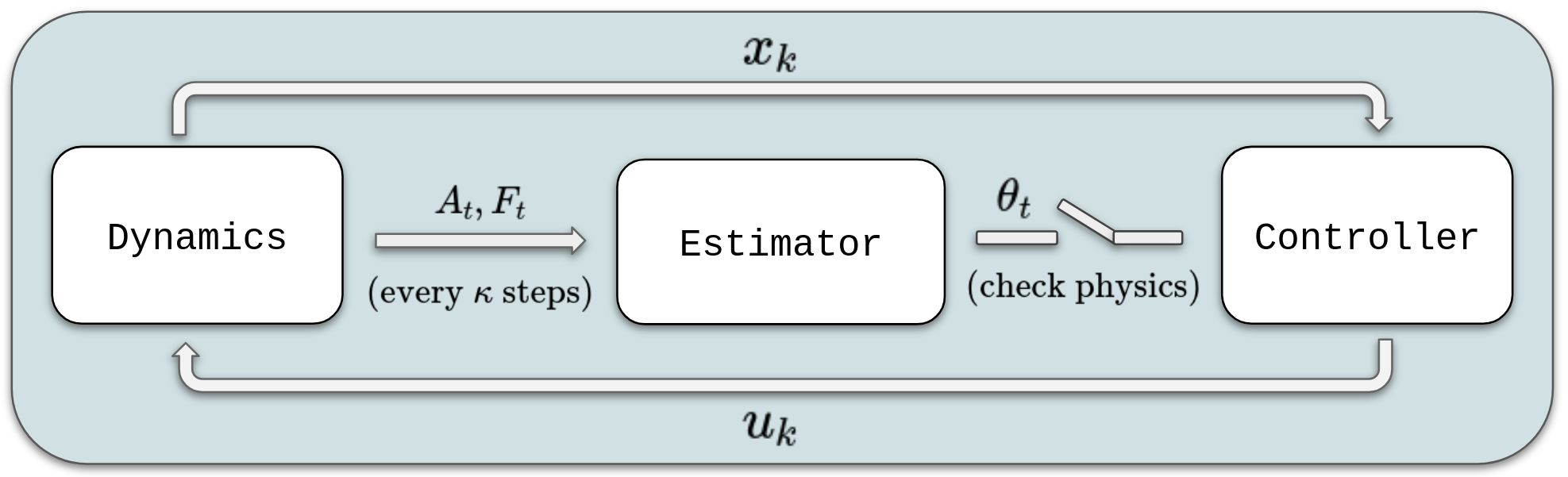}
    \caption{\small \textbf{Online Parameter Estimation Overview}. Closed-loop estimation with fast control updates and slower parameter adaptation. The dynamics model maps control inputs $\mathbf{u}_k$ to states $\mathbf{x}_k$, which generate subsequent controls. Every $\kappa$ steps, the estimator updates controller parameters $\boldsymbol{\theta}_t$ via~\eqref{eq:ssm}. A physics-based safety check~\eqref{eq:safety-filter} safeguards adaptation.}
    \label{fig:estimation_loop}
    \vspace{-15pt}
\end{figure}

For the remainder of the paper, we embed the online parameter estimation problem into the following stochastic state-space framework:
\begin{equation}
\begin{aligned}
    \mathbf{x}_{k+1} &= f(\mathbf{x}_k, \mathbf{u}_k; \boldsymbol{\theta}_t) + \boldsymbol{\omega}_k, 
    &\quad \boldsymbol{\omega}_k &\sim \mathcal{N}(0, \mathbf{R}), \\
    \mathbf{F}_t &= \mathbf{A}_t \boldsymbol{\theta}_t + \nu_t,  &\quad \boldsymbol{\nu}_t &\sim \mathcal{N}(0, \mathbf{Q}). 
\end{aligned}
\label{eq:ssm}
\end{equation}
where $\mathbf{x}_k$ and $\mathbf{u}_k$ denote the robot state and control input, and $\omega_k, \nu_k$ denote the process and measurement noise. We distinguish two time indices: a fast control loop indexed by $k$, and a slower estimation loop indexed by $t$, where parameters $\boldsymbol{\theta}_t$ are updated every $\kappa$ control steps (Fig.~\ref{fig:estimation_loop}).

Applying a forward Euler approximation, the discrete-time dynamics model is defined as follows: 
\begin{equation} 
f(\mathbf{x}_k, \mathbf{u}_k; \boldsymbol{\theta}) = \begin{bmatrix} \mathbf{q}_k + \Delta t \, \dot{\mathbf{q}}_k \\ \dot{\mathbf{q}}_k + \Delta t \, \ddot{\mathbf{q}}_k \end{bmatrix} 
\end{equation}
with the online parameter estimation cast as a weighted least-squares problem: \begin{equation}
\hat{\boldsymbol{\theta}}_t = \arg\min_{\boldsymbol{\theta}_t} \| \mathbf{F}_t - \mathbf{A}_t \boldsymbol{\theta}_t \|_2^2. 
\label{eq:est_obj}
\end{equation} 

The central challenge, therefore, is to update $\hat{\boldsymbol{\theta}}_t$ in a manner that is \emph{efficient}, running in real time on resource-constrained platforms; \emph{rapid}, adapting to abrupt parameter changes during operation; and \emph{stable}, remaining robust under diverse real-world noise conditions.

\subsection{Traditional Online Parameter Estimation Methods} 
\subsubsection{Recursive Least Squares (RLS)}  
RLS computes parameter updates by minimizing a cumulative squared prediction error and then updating the inverse system covariance via the matrix inversion lemma. Modern variants attempt to improve robustness and flexibility: Recursive Total Least Squares (RTLS) mitigates bias from noisy regressors~\cite{RTLS,RGTLS}, variable-forgetting RLS adapts weighting to track varying parameters~\cite{rls_variable_forgetting,1468506}, and kernelized RLS maps the regression into nonlinear feature spaces~\cite{rls_kernel}. Despite these extensions, a fundamental disadvantage of RLS remains: the forgetting factor must be constrained to a narrow, hand-tuned memory window, which in practice lies extremely close to one~\cite{1468506,rls_variable_forgetting}. This makes the method poorly suited to handling abrupt shifts, since smaller forgetting factors cause instability while larger ones prevent adaptation. As a result, adaptive RLS variants remain not adaptive enough for versatile online estimation.

\subsubsection{The Kalman Filter (KF)}  
KF interprets parameters as stochastic states and computes the gain by balancing prior uncertainty against measurement noise. Extensions include: adaptive covariance adjustment~\cite{2017_Adaptive_covariance_adjustment}, Ensemble Kalman Filters (EnKF) for high-dimensional nonlinear systems~\cite{EnselmbleKF}, and square-root formulations for numerical stability~\cite{SQRT_KF,SQRT_UKF}. While such methods improve robustness, they introduce significant computational burden and remain sensitive to initialization and noise-model assumptions, making them less practical for embedded online estimation.

\subsection{Randomized Kaczmarz Methods:} 
The Randomized Kaczmarz (RK) algorithm~\cite{rk} builds upon the classical iterative Kaczmarz method \cite{ferreira2024surveyclassiterativerowaction}. It solves a linear system $\mathbf{A}\boldsymbol{\theta}=\mathbf{b}$ by randomly selecting a row $\mathbf{a}_i^\top$ with probability proportional to $\|\mathbf{a}_i\|_2^2$ and then orthogonally projecting the current estimate onto the row's hyperplane $\{\boldsymbol{\theta}:\langle \mathbf{a}_i,\boldsymbol{\theta}\rangle=b_i\}$, yielding an expected exponential rate of convergence for consistent linear systems~\cite{rk}.
The iterative update is given by:
\begin{equation}
\boldsymbol{\theta}_{t+1}
= \boldsymbol{\theta}_{t}
  + \frac{r_i}{\|\mathbf{a}_i\|_2^{2}}\,\mathbf{a}_i, \quad r_i = b_i - \langle \mathbf{a}_i, \boldsymbol{\theta}_{t} \rangle.
\label{eq:rk-update}
\end{equation}

Several recent variants of these methods have emerged to improve convergence and robustness. 
Greedy Randomized Kaczmarz (GRK)~\cite{bai2018greedy} biases selection toward rows with larger residuals, prioritizing directions where the estimate is most inconsistent with the data and converging at a faster exponential rate than RK for consistent linear systems.
Tail-Averaged Randomized Kaczmarz (TARK)~\cite{epperly2025randomizedkaczmarztailaveraging} further stabilizes performance by averaging recent iterates, smoothing fluctuations and converging at a hybrid rate that balances exponential and polynomial convergence for inconsistent linear systems arising from noise or unmodeled dynamics. 
Together, these mechanisms extend RK into regimes relevant for online parameter estimation.
\section{Recursive Estimation from RLS and KF to Kaczmarz Algorithms} \label{sec:background}
To provide further motivation before explaining our approach, we look at online parameter estimation from a unified recursive–gain perspective. Recursive algorithms can be expressed in the generic form:
\begin{equation}
    \hat{\boldsymbol{\theta}}_{t+1} = \hat{\boldsymbol{\theta}}_{t} + \mathbf{K}_t \boldsymbol{\varepsilon}_t ,
\label{eq:recursive_gain}
\end{equation}
where $\hat{\boldsymbol{\theta}}_t$ is the current estimate, $\boldsymbol{\varepsilon}_t = \mathbf{b}_t - \mathbf{A}_t \hat{\boldsymbol{\theta}}_t$ is the prediction error, and $\mathbf{K}_t$ is the gain matrix that determines how new data influences each update. The distinction among methods lies not in this form, but in how $\mathbf{K}_t$ is constructed, directly impacting computational cost and adaptivity.

\begin{table}[t]
    \centering
    \caption{\small \textbf{Comparison of Parameter Estimation Methods}. Columns summarize desirable properties of online estimators, and rows list representative approaches. TAG-K satisfies all listed criteria while maintaining linear-time complexity per iteration.}
    \label{tab:capability}
    \setlength{\tabcolsep}{3pt} 
    \resizebox{\columnwidth}{!}{
    \begin{tabular}{lccccc}
        \toprule
        \multirow{3}{*}{\textbf{Methods}} & \multicolumn{5}{c}{\textbf{Features}}\\
        \cmidrule{2-6}
        \multirow{2}{*} & {Low} & {Robust to} & {Handles Fast} & {Physics} & {Computational}\\
        & {Cost} & {Process Noise} & {Changes} & {-Reliable} & {Complexity }\\
        \midrule
        RLS & \redcross & \redcross & \redcross & \greencheck &$O(m^3)$  \\
        \hdashline
        KF  & \redcross & \greencheck & \redcross & \greencheck & $O(m^3)$\\
        \hdashline
        Learning & \redcross  & \greencheck & \greencheck & \redcross & - \\
        \hdashline
        \textbf{TAG-K} & \greencheck & \greencheck & \greencheck & \greencheck & $T\cdot O(mn)$\\
        \bottomrule
    \end{tabular}
    }
    \vspace{-15pt}
\end{table}

RLS constructs $\mathbf{K}_t$ from recursive covariance minimization, with the
forgetting factor $\lambda \in (0,1]$ and where $\mathbf{A}_t \in \mathbb{R}^{m\times n}$ denotes the block of $m$ regressors at time $t$:
\begin{equation}
\begin{split}
\mathbf{K}_t^{\text{RLS}} &= \mathbf{P}_{t-1} \mathbf{A}_t^\top (\lambda \mathbf{I} + \mathbf{A}_t \mathbf{P}_{t-1} \mathbf{A}_t^\top)^{-1},\\
\mathbf{P}_t &= \lambda^{-1}\!\left(\mathbf{I} - \mathbf{K}_t^{\text{RLS}} \mathbf{A}_t\right)\mathbf{P}_{t-1}.
\end{split}
\end{equation}
 
Because we operate in a block-update setting that naturally comes with online parameter estimation, each step requires a matrix inversion of size $m$, leading to an $O(m^3)$ time complexity (as noted in Table~\ref{tab:capability})~\cite{sayed2008adaptive}.  

KF, on the other hand, interprets parameters as stochastic states, propagating and updating error covariances with block measurements $\mathbf{A}_t \in \mathbb{R}^{m\times n}$. In this case, the gain matrix is:
\begin{equation}
\mathbf{K}_t^{\text{KF}} = \mathbf{P}_{t|t-1}\mathbf{A}_t^\top\!\left(\mathbf{A}_t \mathbf{P}_{t|t-1}\mathbf{A}_t^\top + \mathbf{R}_t\right)^{-1},
\end{equation}
where, due to the block formulation, each update requires inversion of an $m \times m$ matrix, incurring $O(m^3)$ complexity per step (again noted in Table~\ref{tab:capability})~\cite{kalman1960new}. While optimal under linear–Gaussian assumptions, KF variants remain expensive and fragile under non-stationary conditions.

Finally, Kaczmarz updates also take the form of Eq.~\eqref{eq:recursive_gain}, but now each update is based on a single row $a_i^\top$. The prediction error $\varepsilon_t$ reduces to a scalar residual as in Eq.~\eqref{eq:rk-update}, and the corresponding gain is the row-dependent vector
\begin{equation}
\mathbf{K}_t^{\mathrm{RK}} = \frac{\mathbf{a}_i}{\|\mathbf{a}_i\|_2^{2}},
\label{eq:rk-gain}
\end{equation}

Unlike RLS and KF, which operate on the full regressor matrix at each update, require only $T \le m$ iterations in practice and need not traverse the entire row space. As each TAG-K iteration is dominated by the greedy selection, at $O(mn)$ time per update, a sweep of $T$ rows requires $T\cdot O(mn)$ operations (see Table~\ref{tab:capability})~\cite{bai2018greedy,rk}, making it particularly attractive for embedded, resource-constrained robots.
Moreover, online parameter estimation typically yields overdetermined systems for which Kaczmarz methods remain fast and accurate~\cite{Zouzias_2013, ferreira2024surveyclassiterativerowaction, rk}, as new equations accumulate faster than parameters are updated.
\section{Tail-Averaged Greedy Kaczmarz} \label{sec:method}
In this section, we introduce the Tail-Averaged Greedy Kaczmarz (TAG-K) algorithm, a Kaczmarz variant tailored for online parameter estimation for computationally constrained robotic systems. TAG-K combines greedy randomized row selection with tail averaging, enabling rapid closed-loop adaptation, robustness to measurement and process noise, and lightweight computational complexity. This combination makes it well-suited for real-time robotic applications where both speed and reliability are essential.

The foundation of TAG-K lies in the broader Kaczmarz family of algorithms, which update solutions through projection-based steps that sequentially assimilate measurements. Compared to Recursive Least Squares (RLS) and Kalman Filters (KF), which often suffer from higher computational cost or instability, Randomized Kaczmarz (RK) and its variants provide adaptive and efficient updates. RK achieves exponential convergence in low-noise regimes under standard row-norm sampling assumptions~\cite{rk,Zouzias_2013}. In noisier or streaming settings, its randomized sampling can naturally emphasize \emph{more informative} constraints: in the standard scheme, row $i$ is selected with probability $p_i=\|a_i\|_2^2/\|A\|_F^2$, so higher-norm (higher-energy) measurements are enforced more frequently, which can empirically accelerate adaptation~\cite{rk,Zouzias_2013}. Moreover, in the presence of measurement noise or inconsistency, RK is known to converge to an error neighborhood whose radius depends on the noise level~\cite{needell2010randomized}. TAG-K further extends these benefits by adapting the greedy row selection in GRK~\cite{bai2018greedy} and the tail averaging in TARK~\cite{epperly2025randomizedkaczmarztailaveraging} to achieve both faster convergence and improved robustness. As summarized in Table~\ref{tab:capability}, TAG-K outperforms traditional RLS, KF, and learning-based methods across several axes. It offers low computational cost, robustness to noise, responsiveness to sudden system changes, and physical reliability, all while maintaining the lowest-order computational complexity.

\begin{algorithm}[t!]
\begin{spacing}{1.25}
\caption{TAG-K ($\mathbf{A}, \mathbf{b}, \hat{\boldsymbol{\theta}}_0, T, t_b, \hat{\boldsymbol{\theta}}_{s}=\mathbf{0}\;\;\rightarrow\;\; \hat{\boldsymbol{\theta}}_{t+1}$)}
\label{alg:tagk}
\begin{algorithmic}[1]
    \For{$t = 0$ in $\dots T$}
        \State\textit{// Greedy Row Selection}
        \State $\epsilon_t, \tau_t \gets$ Compute greedy candidates via (\ref{eq:threshold}),  (\ref{eq:greedy_selection})
        \State $i_t \gets$ Sample via (\ref{eq:greedy_prob})
        \State\textit{// Online Update}
        \State $\hat{\boldsymbol{\theta}}_{t+1} (i_t) \gets$ KZ Gain update via (\ref{eq:rk-update}), (\ref{eq:rk-gain})
        \State\textit{// Tail Averaging}
        \If{$t \geq t_b$}
            \State $\hat{\boldsymbol{\theta}}_{s} \gets \hat{\boldsymbol{\theta}}_{s} + \hat{\boldsymbol{\theta}}_{t+1}$
        \EndIf
    \EndFor
    \State $\hat{\boldsymbol{\theta}}_{avg} \gets$ Tail-averaging via (\ref{eq:tail_avg})
    \State \Return $\hat{\boldsymbol{\theta}}_{avg}$
\end{algorithmic}
\end{spacing}
\end{algorithm}

Before performing the Kaczmarz gain update rule defined in~\eqref{eq:rk-update}, we select the target row(s) through a \emph{greedy selection} step which prioritizes rows that contribute most to the current residual error.  
First, a threshold parameter \(\epsilon_k\) is computed to adaptively scale residual importance:
\begin{equation}
    \epsilon_t = \tfrac12 \left(
    \frac{\max_i (r_i^2 / s_i)}{R^2} + \frac{1}{\|A\|_F^2}
    \right),
    \label{eq:threshold}
\end{equation}
where \(s_i = \|A^{(i)}\|_2^2\) and \(R = \|r\|_2\). Then, a candidate set of indices with the largest scaled errors is formed,
\begin{equation}
    \tau_t = \{\, i \mid r_i \ge \epsilon_t R s_i \,\},
    \label{eq:greedy_selection}
\end{equation}
where finally, row $i_t$ is sampled with probability
\begin{equation}
    \mathbb{P}(i) = \frac{r_i^2}{\sum_{j \in \tau_t} r_j^2}, \quad i \in \tau_t .
    \label{eq:greedy_prob}
\end{equation}

We modify the gain update~\eqref{eq:rk-gain}, by incorporating these new sampling probabilities~\eqref{eq:greedy_prob} instead of using probabilities weighted by row norms as seen in the classical RK formulation.
This step biases the gain such that it updates towards directions with higher instantaneous error, enabling faster convergence. 
In online estimation, such targeted updates allow rapid adaptation from only a short data window, which is essential for closed-loop robotic control.

While greedy selection improves adaptation speed, it can also increase estimator variance in the presence of noise or inconsistency. To mitigate this, TAG-K employs \emph{tail averaging}~\cite{epperly2025randomizedkaczmarztailaveraging}: after a burn-in period $t_b$, the final estimate is formed as the average of the most recent iterates,
\begin{equation}
    \hat{\theta}_{avg} = \frac{1}{t-t_b} \sum_{s=t_b}^t \hat{\theta}_s ,
    \label{eq:tail_avg}
\end{equation}
which smooths fluctuations and guarantees convergence to a least-squares solution even in inconsistent systems~\cite{epperly2025randomizedkaczmarztailaveraging}. For robotics applications, where process and sensor noise are unavoidable, this averaging provides a critical safeguard against estimator instability.
\section{Experiments and Results} \label{sec:results}
We evaluate TAG-K against a suite of baselines to assess parameter-estimation accuracy, computational runtime, and downstream robotic performance. Specifically, we aim to answer the following questions:  
\begin{enumerate}
    \item[\textbf{Q1:}] Does TAG-K produce faster and more accurate parameter estimates than existing baselines?  
    \item[\textbf{Q2:}] Which features of estimation performance or algorithmic design most strongly influence performance? 
    \item[\textbf{Q3:}] Does TAG-K reduce on-device compute compared to baselines on resource-constrained hardware?
\end{enumerate}

To address these questions, we benchmark TAG-K and baselines on online inertial parameter estimation in simulation (\ref{sec:e2e_perf}), study the contribution of key design choices and estimation-time behavior via targeted analyses and ablations (\ref{sec:early_adapt}, \ref{sec:ablation}), and measure runtime and resource usage on microcontrollers (\ref{sec:micro_controller}).

\subsection{Quadrotor Under Unknown Payload} \label{sec:exp:vs_baselines}
\subsubsection{Experiment Setup}
We first perform a three-part evaluation of our approach leveraging a simulated model of a Crazyflie 2.1~\cite{crazyflie2_1} quadrotor, a widely used testbed in edge robotics research~\cite{crazyswarm,crazychoir,rlcrazyflie,tinympc}, and evaluate its performance under unknown payload events to address \textbf{Q1} and \textbf{Q2}. These experiments were conducted on a laptop with a 3.3 GHz AMD Ryzen 9 5900HS CPU.

In each experiment, the quadrotor starts at a randomized location, and tracks one of five reference trajectories: a Figure-8, Circle, Spiral, Helix, or Ellipse. Across each, the peak linear and angular speeds are limited to \(0.5\,\mathrm{m/s}\) and \(0.5\,\mathrm{rad/s}\), respectively. For stabilization and trajectory tracking, we employ a standard LQR controller~\cite{lewis12optimal} running at \(50\,\mathrm{Hz}\), while the parameter estimator runs at \(2.5\,\mathrm{Hz}\) using a 5-frame history of the 10 inertial parameters (the body mass plus $3\times3$ inertia-matrix components). The estimator minimizes the weighted least-squares loss in~\eqref{eq:est_obj}, while the LQR controller minimizes the mean squared tracking error $\|e\|_2^2$. Each experiment lasts \(20\,\mathrm{s}\). To emulate a payload-transfer task, a payload equal to \(30\text{–}50\%\) of the vehicle mass is attached at an offset of \(20\text{–}30\%\) of the arm length at a random time in \([4,6]\,\mathrm{s}\) and removed at a random time in \([12,14]\,\mathrm{s}\). At each parameter-estimation step, we re-linearize the dynamics about the current operating point so that updated inertial parameters are consistently reflected in the control-affine model.

To model real-world sensor imperfections, we inject Gaussian measurement noise at four levels, ranging from noise-free to a maximum standard deviation of $2.5\,\mathrm{cm/s}$ in velocity and $0.25\,\mathrm{cm/s^2}$ in acceleration, applied to the state inputs of both controller and estimator. For baselines, we conduct a grid search over key hyperparameters (e.g., forgetting factor, initial covariance magnitude, Kalman-filter noise priors) and report the best configuration for each method under the end-to-end tracking objective~\cite{m2015introductionkalmanfiltertuning, 6288731}. 
To prevent catastrophic outcomes for \emph{baseline} methods, we use a lightweight inertial-parameter safety filter that rejects physically invalid estimates, where $L_{\mathrm{arm}}=0.0325$ and $\alpha=10^{-4}$:
\begin{equation}\label{eq:safety-filter}
\left\{\, m>0,\;\; \|\mathbf{r}_{\mathrm{CoM}}\|\le L_{\mathrm{arm}},\;\; I\succ 0,\;\; \lambda_{\max}(I) \le \alpha \;\right\}
\end{equation}

\begin{figure*}[t]
    \centering
    \includegraphics[width=0.99\textwidth]{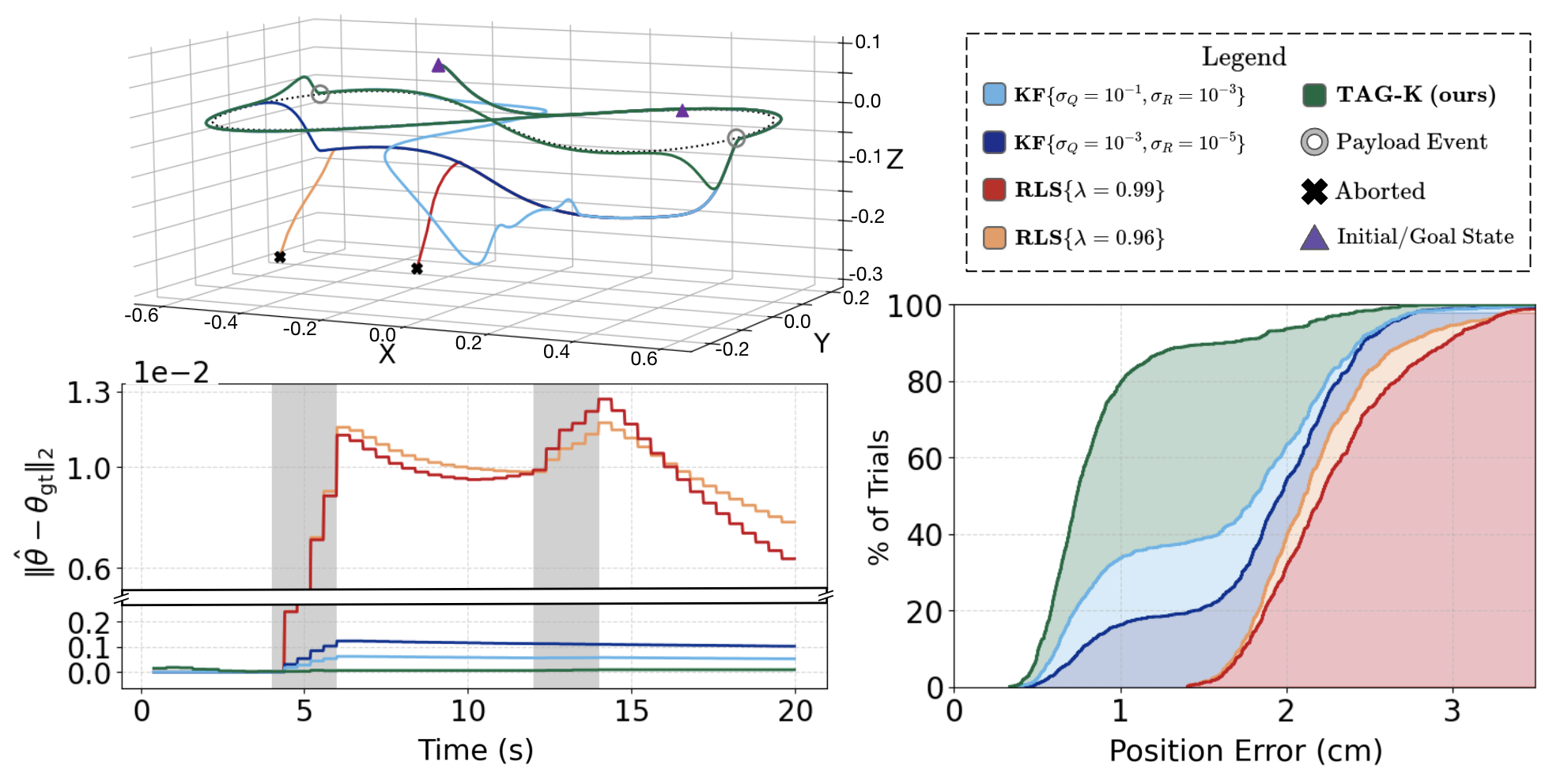}
    \captionsetup{width=\textwidth}
    \caption{\small \textbf{Tracking and Estimation Performance}. \emph{Upper Left}: Example single trial of a quadrotor tracking a figure-8 reference trajectory (dotted line, units in meters) with unknown payload add/drop events (gray rings).
    \emph{Lower Left}: Prediction error timeseries comparing TAG-K with baseline estimators averaged over 2,000 trials per estimator. Grey shaded regions denote payload add/drop events, highlighting TAG-K’s rapid and consistent re-convergence after disturbances.
    \emph{Lower Right}: Cumulative distribution function (CDF) of position tracking error over 2,000 trials per estimator, averaged across four different noise levels. TAG-K achieves the lowest overall error, demonstrating superior adaptation and end-to-end tracking performance.}
    \label{fig:combined}
    \vspace{-10pt}
\end{figure*}
\begin{figure*}[t]
    \centering
    \begin{subfigure}[t]{0.43\textwidth}
        \includegraphics[width=0.99\linewidth]{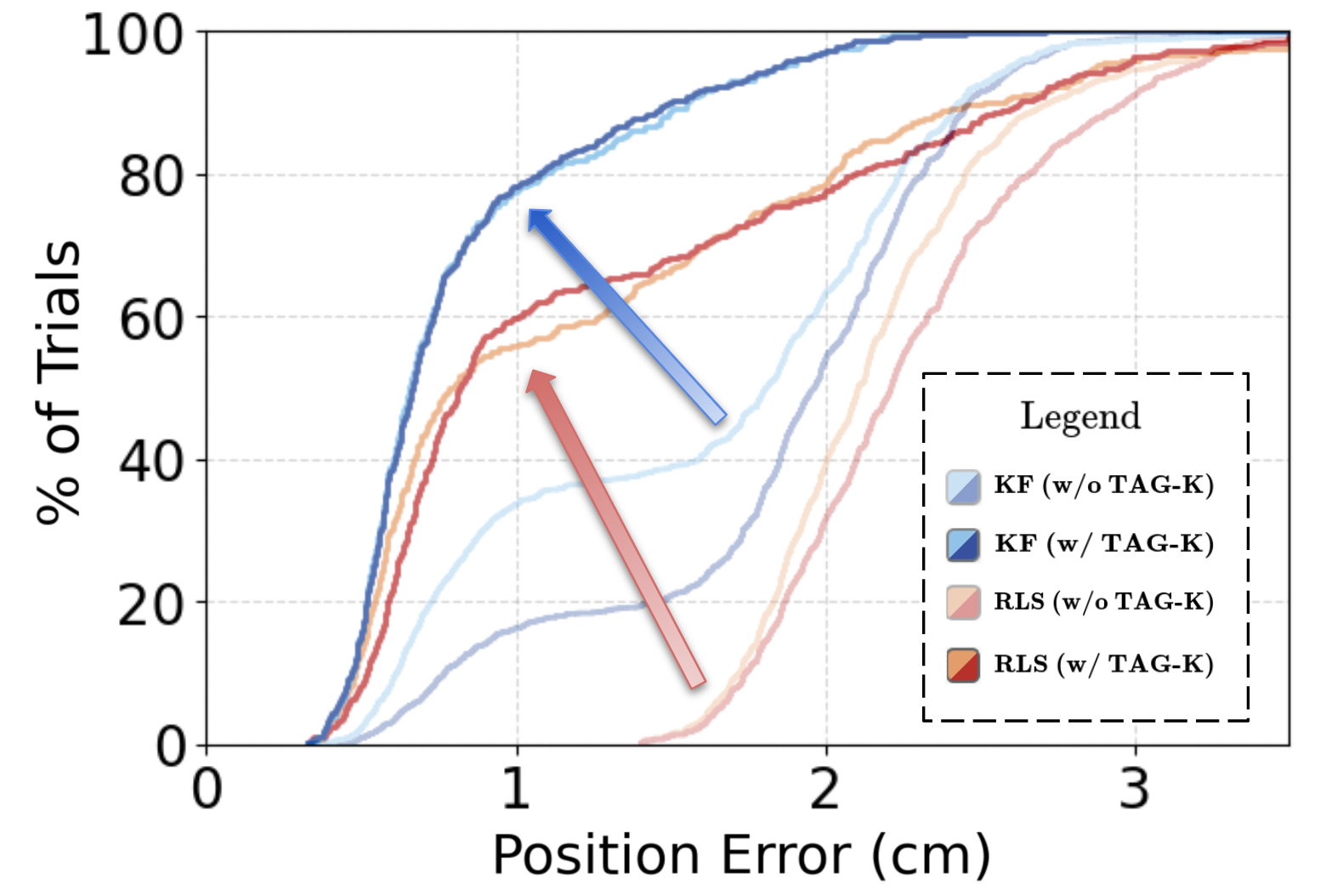}
        \caption{Position error CDF before (light) and after (dark) substituting the first post-event estimate with TAG-K’s estimate.}
        \label{fig:exp2_ecdf_ts}
    \end{subfigure}
    \hfill
    \begin{subfigure}[t]{0.55\textwidth}
        \includegraphics[width=0.99\linewidth]{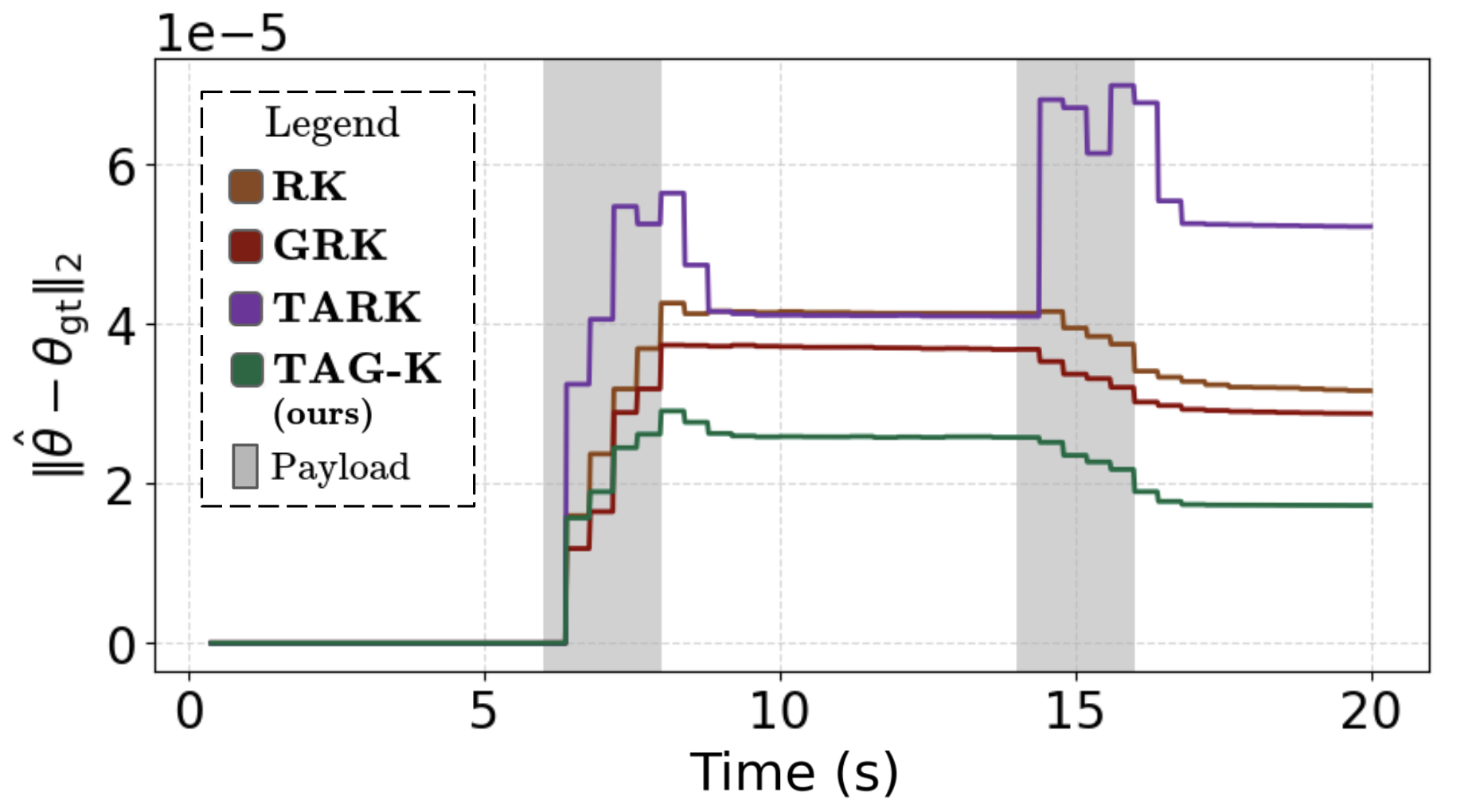}
        \caption{Prediction error comparison across Kaczmarz variants. Grey regions indicate payload addition/removal events.}
        \label{fig:exp3_theta_err_ts}
    \end{subfigure}
    \caption{\small \textbf{Ablation of Early Adaptation and Algorithmic Design}. 
    Left: Substituting the first post-event estimate from baseline estimators with TAG-K’s estimate improves tracking error, highlighting the importance of rapid early adaptation. 
    Right: Comparison with Kaczmarz variants shows that combining greedy row selection and tail averaging yields superior robustness and adaptation under abrupt payload changes (grey regions).}
    \vspace{-15pt}
\end{figure*}
\begin{table*}[t]
  \centering
  \caption{\small \textbf{Quadrotor Tracking under Varying Measurement Noise}. We run 500 trials per noise level per algorithm across five reference trajectories with randomized initial states, trajectory parameters, event times, and payload mass/location. RLS-low/high is $\lambda{=}0.99/0.96$; KF-low/high is $\sigma_Q{=}10^{-3}/10^{-1},\ \sigma_R{=}10^{-5}/10^{-3}$. $(t{+}1)$ Est. Error is the mean estimation error one step after the payload event.}
  \label{tab:algo-summary-noise}
  \small
  \setlength{\tabcolsep}{4pt}
  \renewcommand{\arraystretch}{1.15}
  \begin{tabular}{llccccccc}
    \toprule
    & & \multicolumn{3}{c}{\textbf{Estimation Accuracy}} & \multicolumn{2}{c}{\textbf{Tracking Performance}} & \multicolumn{2}{c}{\textbf{Iteration Time ($\boldsymbol{\mu}$s)}} \\
    \midrule
    \textbf{Noise} & \textbf{Algorithm} & \textbf{Pos. Error (cm)} & \textbf{Mean Est. Error} & \textbf{($\mathbf{t+1}$) Est. Error} & \textbf{Success} & \textbf{Aborted} & \textbf{Median} & \textbf{P95} \\
    \midrule

    \multirow{5}{*}{\textbf{none}}
      & RLS-low          & 2.13 & 8.47e-03 & 1.10e-02 & 0.0\% & 11.8\% & 63.1 & 92.1 \\
      & RLS-high            & 2.18 & 8.10e-03 & 1.00e-02 & 0.0\% & 22.2\% & 64.4 & 90.0\\
      & KF-low            & 1.79 & 1.10e-03 & 1.42e-03 & 20.2\% & 4.4\% & 74.1 & 94.0\\
      & KF-high           & 1.40 & 4.62e-04 & 5.99e-04 & 48.2\% & 3.6\% & 73.9 & 97.8\\
      & \textbf{TAG-K (ours)} & \textbf{0.84} & \textbf{3.92e-04} & \textbf{7.11e-05} & \textbf{87.8\%} & \textbf{3.6\%} & \textbf{40.4} & \textbf{59.6} \\
    \midrule

    \multirow{5}{*}{\textbf{low}}
      & RLS-low          & 2.13 & 8.45e-03 & 1.10e-02 & 0.0\% & 12.0\% & 62.5 & 94.5 \\
      & RLS-high           & 2.18 & 8.08e-03 & 1.00e-02 & 0.0\% & 22.0\% & 64.2 & 91.4 \\
      & KF-low           & 1.78 & 1.09e-03 & 1.39e-03 & 21.6\% & 3.6\% & 74.3 & 96.2\\
      & KF-high           & 1.40 & 4.61e-04 & 5.97e-04 & 49.2\% & \textbf{3.2\%} & 74.9 & 101\\
      & \textbf{TAG-K (ours)} & \textbf{0.83} & \textbf{3.14e-04} & \textbf{7.06e-05} & \textbf{87.8\%} & 3.6\% & \textbf{43.0} & \textbf{60.3} \\
    \midrule

    \multirow{5}{*}{\textbf{medium}}
      & RLS-low          & 2.18 & 7.47e-03 & 8.99e-03 & 0.2\% & 17.8\% & 63.7 & 94.0\\
      & RLS-high           & 2.30 & 6.63e-03 & 8.14e-03 & 0.2\% & 39.4\% & 66.2 & 93.7 \\
      & KF-low           & 1.76 & 7.70e-04 & 9.26e-04 & 28.4\% & 5.4\% & 75.1 & 113 \\
      & KF-high           & 1.56 & 4.42e-04 & 5.60e-04 & 43.6\% & \textbf{3.8\%} & 75.6 & 118 \\
      & \textbf{TAG-K (ours)} & \textbf{0.84} & \textbf{3.12e-04} & \textbf{9.89e-05} & \textbf{87.4\%} & 4.0\% &  \textbf{43.7} & \textbf{60.9} \\
    \midrule

    \multirow{5}{*}{\textbf{high}}
      & RLS-low      & 2.25 & 5.75e-03 & 6.24e-03 & 0.6\% & 23.0\% & 63.5 & 94.4 \\
      & RLS-high           & 2.47 & 5.00e-03 & 5.46e-03 & 0.4\% & 49.4\% & 63.2 & 90.8 \\
      & KF-low         & 2.05 & 4.47e-04 & 4.84e-04 & 17.4\% & \textbf{15.0\%} & 73.9 & 96.5 \\
      & KF-high           & 2.05 & 3.72e-04 & 4.06e-04 & 22.0\% & 17.0\% & 73.5 & 97.8 \\
      & \textbf{TAG-K (ours)} & \textbf{1.03} & \textbf{2.86e-04} & \textbf{8.46e-05} & \textbf{58.4\%} & 31.4\% & \textbf{42.5} & \textbf{58.2} \\
    \bottomrule
  \end{tabular}
  \vspace{-10pt}
\end{table*}

\subsubsection{Estimation and End-to-End Performance}
\label{sec:e2e_perf}
We visualize a representative trial from this experiment set in Figure~\ref{fig:combined} (upper left). In this trial, each estimator is initialized from the same start state and tracks the dotted figure-8 reference toward a pre-set goal state (both marked with purple triangles) under injected measurement noise using the aforementioned LQR controller. Payload add/drop events are marked with gray rings. After each event, TAG-K, by its greedy mechanism, rapidly adapts its inertial parameters and re-converges to the reference, whereas baselines mis-estimate the parameters, drift persistently, and in several cases abort due to compounding errors (marked with black Xs). In this specific trial, only TAG-K completes the task. However, baseline methods can complete the task in many other trials.

Figure~\ref{fig:combined} (bottom) summarizes the aggregated outcomes across all 2{,}000 trials per estimator method. The bottom left panel presents the time evolution of the average parameter estimation error, while the bottom right panel shows the cumulative distribution function (CDF) of the end-to-end tracking position error. Together, these plots highlight the differences in both convergence behavior and final tracking accuracy. We observe that RLS and KF consistently maintain higher parameter estimation errors, which directly translate into larger tracking deviations. By contrast, TAG-K rapidly adapts after payload events, quickly reducing estimation error and maintaining accurate parameters. This fast and stable adaptation yields markedly lower tracking errors across the full distribution of trials, demonstrating our method’s ability to robustly handle abrupt system changes.

Table~\ref{tab:algo-summary-noise} provides a more detailed breakdown by noise regime, further reinforcing these observations. Note that we consider a trial aborted if the positional tracking error exceeds $30\,\mathrm{cm}$ and is successful if the error remains below $5\,\mathrm{cm}$ at 10 steps after the payload event, without being previously aborted. We use this success metric to avoid crediting estimators that fail to adapt during the payload addition event but coincidentally exhibit low estimation error after the payload is dropped. 
Across all noise levels, TAG-K consistently achieves the lowest mean estimation and end-to-end position errors, alongside the highest success rates. While KF shows slightly lower abort rates under extreme noise, its diminished success rates reflect closed-loop instability. In contrast, TAG-K maintains stable tracking and accurate parameter estimation even in high-noise conditions. Overall, TAG-K offers the most effective balance of speed, accuracy, and robustness, outperforming all evaluated methods.

\subsubsection{Importance of Early Adaptation}
\label{sec:early_adapt}
Across all trials, inaccurate parameter estimates quickly push the quadrotor into noisier, less stable states, with errors compounding over time. This creates a strong dependence on rapid correction after a disturbance or payload change. We therefore hypothesize that early adaptation, specifically the accuracy of the very first updated estimate, plays a decisive role in overall closed-loop performance. To evaluate the role of the first estimates, we ran controlled trials in which two estimators were executed in parallel. Immediately following a payload event, the baseline estimator’s first update was replaced by TAG-K’s estimate, after which the baseline method continued independently.

As shown in Figure~\ref{fig:exp2_ecdf_ts}, initializing baselines with TAG-K’s first post-event estimate dramatically improves performance, confirming that early correction quality dictates long-term tracking accuracy. These findings are consistent with Table~\ref{tab:algo-summary-noise}, where TAG-K achieves an order-of-magnitude improvement in immediate ($t{+}1$) adaptation. Collectively, these results demonstrate TAG-K’s superior ``step-one corrections,'' underscoring its unique effectiveness in scenarios where rapid and robust adaptation is essential for stable closed-loop operation.

\subsubsection{Kaczmarz Family Ablation}
\label{sec:ablation}
To better understand the impact of greedy selection and tail averaging, we ablate these components by comparing TAG-K to three Kaczmarz variants in Figure~\ref{fig:exp3_theta_err_ts}: 1) Randomized Kaczmarz (RK), which employs random row sampling weighted by row norm without greedy selection or tail averaging; 2) Tail-Averaged Randomized Kaczmarz (TARK), which introduces tail averaging, stabilizing estimates while still retaining row-norm-based random sampling; and 3) Greedy Randomized Kaczmarz (GRK) which uses greedy row selection for faster convergence but lacks the stability provided by averaging.

Across 2{,}000 trials, the results clearly show the value of combining these two changes: reduced estimation errors and iteration times, and improved tracking performance. TAG-K adapts faster than RK and TARK after sudden changes, driven by the selective power of greedy sampling. While GRK alone can sometimes converge marginally faster in the very first iterations, it exhibits higher long-term variability. By incorporating tail averaging, TAG-K reduces this variance and maintains stable estimates, preventing drift and compounding errors in closed-loop operation. This explains TAG-K's closed-loop outperformance, achieving both rapid initial adaptation and strong long-term robustness.

\begin{table*}[t]
\centering
\caption{\small \textbf{Benchmark Runtimes across Parameter Dimensions.} 
Average per-update runtime ($\mu$s) over 10 trials on a Teensy 4.1 microcontroller. Baseline abbreviations follow Table~\ref{tab:algo-summary-noise}. TAG-K achieves 4.7$\times$--20.7$\times$ speedups, with gains increasing at larger dimensions.}
\label{tab:benchmark_results}
\begin{tabular}{lcccccc}
\toprule
\textbf{Parameters} & \textbf{RLS-low} & \textbf{RLS-high} & \textbf{KF-low} & \textbf{KF-high} & \textbf{TAG-K} & \textbf{Speedup vs TAG-K} \\
\midrule
40   & 2,292 & 2,289 & 2,279 & 2,279 & \textbf{480} & 4.7--4.8\(\times\) \\
60   & 4,574 & 4,574 & 4,536 & 4,535 & \textbf{597} & 7.6--7.7\(\times\) \\
80   & 8,023 & 8,025 & 7,928 & 7,929 & \textbf{716} & 11.1--11.2\(\times\) \\
100  & 13,011 & 13,011 & 12,851 & 12,850 & \textbf{837} & 15.4--15.5\(\times\) \\
120  & 19,685 & 19,683 & 19,451 & 19,452 & \textbf{950} & 20.5--20.7\(\times\) \\
\midrule
\textbf{Average} & 9,500 & 9,500 & 9,429 & 9,429 & \textbf{716} & 13.2--13.3\(\times\) \\
\bottomrule
\end{tabular}
\vspace{-10pt}
\end{table*}

\subsection{Microcontroller Timing Benchmark}
\label{sec:micro_controller}
To further assess \textbf{Q3}, the computational performance of our approach, we include an additional benchmark on a Teensy 4.1~\cite{teensy41} microcontroller (MCU) development board.\footnote{The Teensy features an ARM Cortex-M7 microcontroller operating at 600 MHz, with 7.75 MB of flash memory, 512 kB of tightly coupled static RAM, and an additional 512 kB of tightly coupled dynamic RAM. All are orders of magnitude less compute than the CPU used in earlier benchmarks, and a common baseline for tiny robotic tasks~\cite{tinympc}.} This experiment highlights how TAG-K’s computational efficiency enables deployment on resource-constrained edge robotic systems.
In this experiment, we benchmark RLS, KF, and TAG-K across a variety of parameter window sizes ($n=40,60,80,100,120$, corresponding to inertial parameter estimation problems for $4, 6, 8, 10,12$ DoF robots. We set the measurement window size to $m=30$ and use the same forgetting factor, noise covariance and other optimized hyperparameters as in the prior setups.

Table~\ref{tab:benchmark_results} summarizes the average per-update runtimes (in $\mu$s) across 10 independent trials, with the final column reporting the relative speedup of TAG-K over the baselines. The results show that TAG-K consistently and substantially outperforms both RLS and KF in computational efficiency. Depending on the parameter dimension, TAG-K achieves speedups ranging from $4.8\times$ to $20.7\times$, with an average improvement of approximately $12\times$ across all tested scenarios. 
This demonstrates that TAG-K not only offers strong estimation accuracy, but also substantially reduces the computational burden. This is especially important for embedded systems, like our target MCU, where limited processing power, small caches, and constrained register files mean that even modest increases in problem size can lead to large increases in computation time. Most importantly, this level of efficiency is particularly valuable in real-time, resource-constrained robotic platforms, where saving even a few microseconds can have a meaningful impact on overall system performance.
\section{Conclusion and Future Work} \label{sec:conclusion}
We presented TAG-K, a tail-averaged greedy Kaczmarz algorithm tailored for online inertial parameter estimation in robotics. By integrating greedy residual-driven row selection with tail averaging, TAG-K achieves both rapid adaptation to abrupt parameter changes and stable convergence under noise. Our experimental evaluations demonstrate that TAG-K consistently outperforms RLS, KF, and existing Kaczmarz variants, yielding up to $2\times$ improvement in end-to-end tracking accuracy, an order-of-magnitude reduction in early post-event estimation error, robust closed-loop performance across diverse trajectories and noise conditions, and up to a $20\times$ reduction in latency on MCU-class hardware.

While our evaluation considered diverse trajectories, noise models, and payload variations in high-fidelity simulation, the next step is validating TAG-K on physical quadrotors to assess robustness under sensing imperfections, delays, and unmodeled disturbances. Beyond quadrotors, TAG-K naturally extends to domains like manipulation, where online inertial and contact estimation is key, and legged locomotion, where adaptive identification of payloads and terrain can improve stability. Finally, future work includes deeper theoretical analysis of convergence and closed-loop stability under biased noise or delays. Together, these directions position TAG-K as a practical framework to enable online parameter estimation for high-performance control across real-world robotic platforms.
\bibliographystyle{IEEEtran}
\bibliography{IEEEabrv,refs.bib}

\end{document}